\documentclass{article}
\usepackage{spconf,amsmath,graphicx}
\usepackage{mathtools}
\usepackage[dvipsnames]{xcolor}
\usepackage{pifont}
\usepackage{soul}
\usepackage[color=true, onerror=nice]{yagusylo}
\defdingname[fourier][global]{116}{rhand}[RubineRed]
\usepackage{booktabs}
\usepackage[pagebackref=true,breaklinks=true,colorlinks,bookmarks=false]{hyperref}

\newcommand{\bslcorpus}{\textsc{BslCorpus}}
\newcommand{\bslonek}{BSL-1K}
\usepackage[font={small}]{caption} 

\usepackage{subcaption}
\usepackage{pgfplots}

\usepackage[numbers,sort]{natbib}

\title{Sign Language Segmentation with Temporal Convolutional Networks}
\name{Katrin Renz$^{1,2}$ \, Nicolaj C. Stache$^{2}$ \, Samuel Albanie$^{1}$ \, G\"ul Varol$^{1,3}$}

\address{$^{1}$ Visual Geometry Group, University of Oxford, UK \\
$^{2}$ University of Heilbronn, Germany\\
$^{3}$ LIGM, \'Ecole des Ponts, Univ Gustave Eiffel, CNRS, France \\
\url{https://www.robots.ox.ac.uk/~vgg/research/signsegmentation/}
}

\begin{document}
%
\maketitle
\begin{abstract}
 The objective of this work is to determine the location of temporal boundaries between signs in continuous sign language videos. Our approach employs 3D convolutional neural network representations with iterative temporal segment refinement to resolve ambiguities between sign boundary cues. We demonstrate the effectiveness of our approach on the \bslcorpus{}, PHOENIX14 and \bslonek{} datasets, showing considerable improvement over the state of the art and the ability to generalise to new signers, languages and domains.

\end{abstract}

\section{Introduction} \label{sec:intro}

Sign languages are languages
that have evolved among deaf communities
that employ movements of the face, body and
hands to convey meaning~\cite{sutton-spence_woll_1999}.
Despite significant recent progress in neural
machine translation for spoken languages~\cite{vaswani2017attention}
and fine-grained visual action recognition~\cite{shao2020finegym},
automatic recognition and translation of sign
languages remains far from human
performance~\cite{koller2020quantitative}.
A key challenge in closing this gap is
the prohibitive annotation cost of
constructing high-quality labelled
sign language corpora, which are consequently
orders of magnitude smaller than their
counterparts in other domains such as
speech recognition~\cite{bragg2019sign}.
The high annotation cost is driven by:
(1) the limited supply of annotators who
possess the skill to annotate the data, and
(2) the laborious time-per-annotation (taking
100 hours to densely label 1 hour of signing
content~\cite{Dreuw2008TowardsAS}).

\begin{figure}
    \centering
    \includegraphics[width=0.48\textwidth]{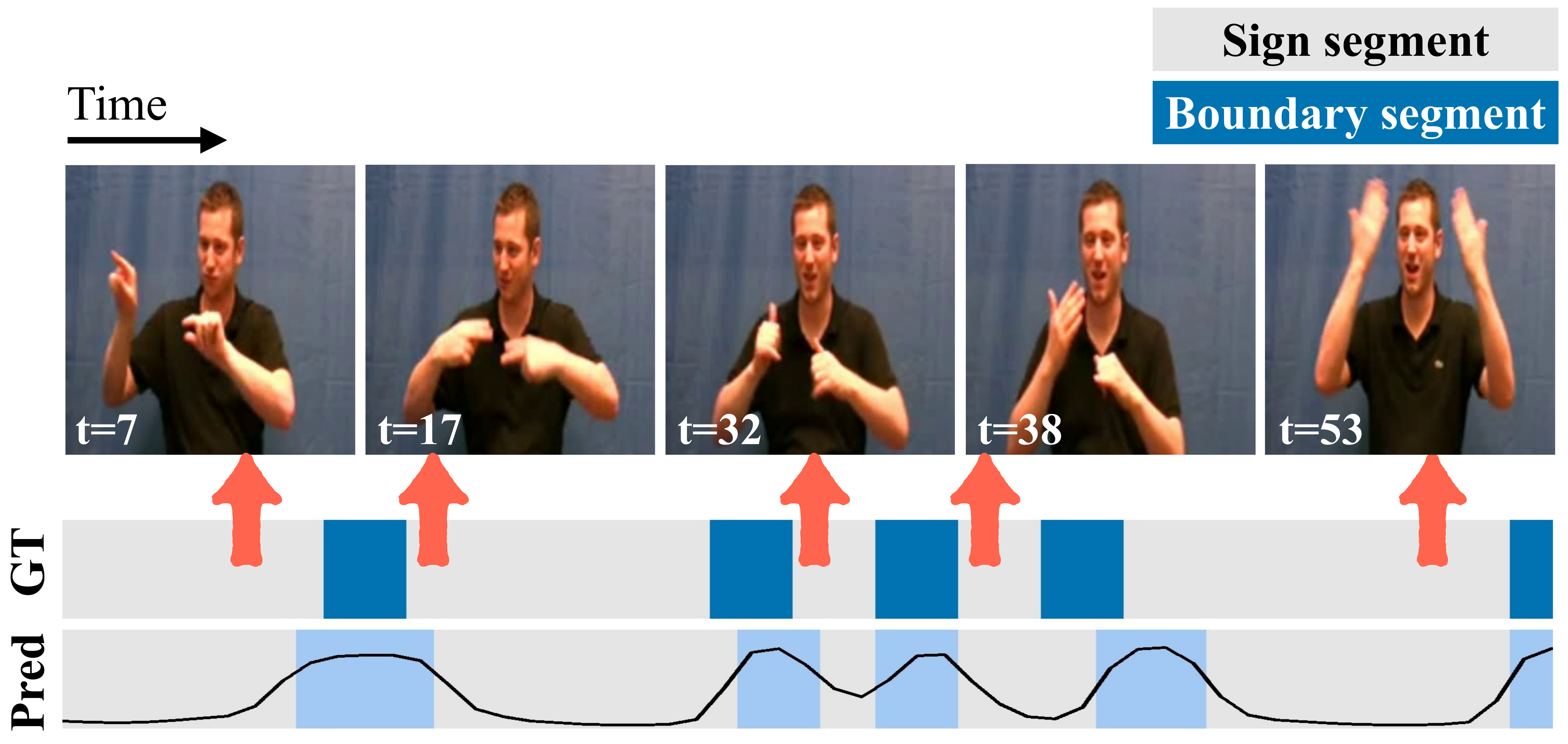}
    \mbox{}\vspace{-0.6cm} \\
    \caption{
    \textbf{Task:} We illustrate the task of temporal sign segmentation for an 
example continuous sign language sequence from \bslcorpus.
    Ground truth (GT) and predictions of our model (Pred)
    are shown, together with a sample frame and its
    frame number from each sign segment
    (whose location is denoted with a red arrow).
    Despite the fast transitions between signs,
    our model is able to accurately detect boundaries.  
    }
    \mbox{}\vspace{-0.6cm} \\
    \label{fig:teaser}
\end{figure}

Motivated by these challenges, the focus of
this work is to propose an automatic sign
segmentation model that can identify the
locations of temporal boundaries between
signs in continuous sign language (see
Fig.~\ref{fig:teaser}).
This task, which has received limited attention in the
literature, has the potential to significantly reduce
annotation cost and expedite novel corpora collection efforts.
Key challenges for such an automatic segmentation
tool include the fast speed of continuous
signing and the presence of motion blur,
especially around hands.

We make the following contributions:
(1) We demonstrate the effectiveness of coupling
robust 3D spatio-temporal convolutional neural network (CNN)
representations with an iterative 1D temporal CNN
refinement module to produce accurate sign boundary
predictions; (2) we provide comprehensive
experiments to study different components
of our method.
(3) We contribute a test set of human-annotated
British Sign Language (BSL) temporal sign segmentation labels
for a portion of \bslonek{} to provide a
benchmark for future work.
(4) We show that our approach strongly outperforms
prior work on the \bslcorpus{} and \bslonek{}
datasets and investigate its cross-lingual
generalisation on PHOENIX14.

\section{Related Work} \label{sec:related}

The linguistic definition of sign boundaries has been non-trivial
in prior work~\cite{brentari2009effects}.
The research of~\cite{braffort2012,gonzalez} showed that the same
signs were annotated differently across teams.
Therefore, several works have attempted standardising the definition of
sign boundaries~\cite{hanke2012,casborn2008,cormierbsl}. The annotation guidelines of \bslcorpus~\cite{cormierbsl}, which we follow, define the position of a boundary at the time when the hands start moving away from the previous sign, an event typically indicated by a change in direction, orientation or handshape.

Although the task of automatically identifying temporal boundaries
between signs has received attention in the literature, it has typically
been tackled with methods that require access to a semantic labelling
of the signed content (e.g., in the form of glosses or free-form sentence translations)~\cite{Santemiz2009,koller2017re,KollerPami20}. 
As we show in Sec.~\ref{sec:experiments}, our approach can
operate effectively with access to only category-agnostic annotation
on the domain of interest. Other works tackled segmenting sign language
content into sentence-like units~\cite{bull2020}, identifying whether a
person is signing or not~\cite{moryossef2020real}, or segmenting signs given subtitles~\cite{Cooper2009LearningSF}. Unsupervised sign sub-unit segmentation has also been explored~\cite{theodorakis2014dynamic}.

Recently,~\cite{farag2019learning} proposed to tackle the category-agnostic
sign boundary segmentation problem using a random forest in combination with
geometric features computed from 3D skeletal information obtained via motion
capture. They demonstrate their approach on a small-scale Japanese Sign Language (JSL) dataset~\cite{Brock2018DeepJA}---we compare our approach with theirs
in Sec.~\ref{subsec:sota}.

\begin{figure}
    \centering
    \includegraphics[width=0.41\textwidth]{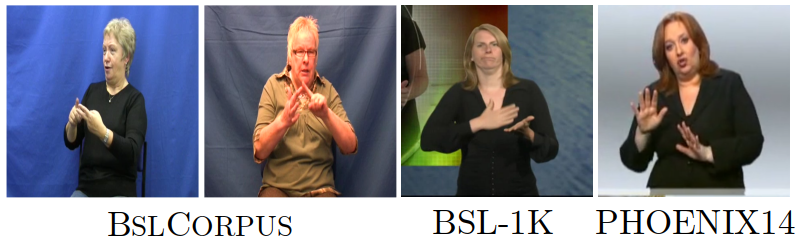}
    \mbox{}\vspace{-0.3cm} \\
    \caption{
    \textbf{Datasets:}
    We provide samples
    from each dataset we use in this work: 
\bslcorpus{}~\cite{bslcorpus17,schembri2013building},
    BSL-1K~\cite{albanie20_bsl1k}, PHOENIX14~\cite{Koller15cslr}.
    }
    \mbox{}\vspace{-0.6cm} \\
    \label{fig:datasets}
\end{figure}
\vspace{-0.2cm}
\section{Sign Segmentation} \label{sec:method}

\noindent \textbf{Problem formulation}.
Given a sequence of video frames of continuous
signing, $\mathbf{x} = (x_1, \dots, x_N)$, the
objective of sign language segmentation is to
predict a corresponding vector of labels
$\mathbf{y} = (y_1, \dots, y_N) \in \{0, 1\}^N$,
where label values of $0$ and $1$ denote the
interior of a sign segment (or ''token'')
and boundaries between sign segments, respectively.
In the sign language corpus construction
literature, several different approaches have been
used to define the precise extent of a sign
token~\cite{hanke2012}. In this work, we follow
the set of conventions for parsing signs prescribed
by~\cite{cormierbsl}.

\noindent \textbf{Training}.
Motivated by its effectiveness for human action recognition
and recently for sign language
recognition~\cite{albanie20_bsl1k,Joze19msasl,li2020word} and sign spotting~\cite{momeni20watchread}, our approach
adopts the spatio-temporal convolutional I3D
architecture~\cite{carreira2017quo} and couples it with
the Multi-Stage Temporal Convolutional Network (MS-TCN)
module proposed by~\cite{Farha_2019_CVPR}.
Each stage of the latter---known as a Single-Stage
TCN (SS-TCN)---comprises a stack of dilated residual
layers of $1$D temporal convolutions, followed by a
linear classifier which is used to predict frame-level
labels, $(y_1, \dots, y_N)$.
Each SS-TCN is run sequentially, such that the predictions
of one stage are used as input to the following stage,
refining the segmentations.
Our model is trained to perform frame-level binary classification
with a cross-entropy loss, together with a smoothing
truncated mean squared loss (following the formulation
described in~\cite{Farha_2019_CVPR}) to reduce over-segmentation errors.
For constructing the ground truth, we assign a video
frame as boundary (i.e., $y=1$) if it is at the start or end time
of the sign segment.
In addition, any frames between the end of one sign
and the start of the next sign are also assigned as boundary.
In Sec.~\ref{sec:experiments}, we conduct an empirical
evaluation of a number of variants of this model,
including: 
the number of stages and 
the importance
of finetuning the I3D backbone.

\vspace{-0.2cm}
\section{Experiments} \label{sec:experiments}

In this section, we describe the datasets used
in our experiments (Sec.~\ref{subsec:datasets}),
present our ablations to assess different
components of our approach (Sec.~\ref{subsec:ablation}),
compare to prior work (Sec.~\ref{subsec:sota}),
test the generalisation capability 
on different datasets (Sec.~\ref{subsec:annot}),
and provide qualitative results (Sec.~\ref{subsec:qual}).


\begin{table}
    \centering
    \resizebox{0.99\linewidth}{!}{
    \begin{tabular}{lc @{\hspace{0.1cm}}lc@{\hspace{0.1cm}}lc@{\hspace{0.1cm}}l}
         \toprule
          & train && val && test \\
         \midrule
         avg \#frames per sign & 11.3 & {\footnotesize $\pm$ 8.5} & 11.3 & {\footnotesize $\pm$ 7.9} & 11.5 & {\footnotesize $\pm$  8.6}\\
         avg \#frames per video & 80.8 & {\footnotesize $\pm$33.3}  & 81.2 & {\footnotesize $\pm$33.3} & 80.8 & {\footnotesize $\pm$32.8} \\
         avg \#glosses per video & 6.8 & {\footnotesize $\pm$ 1.8}  & 7.1 & {\footnotesize $\pm$ 1.9}  &  6.9 & {\footnotesize $\pm$ 1.8} \\
         \midrule
         total \#videos & 5413 && 763 && 703 \\
         total \#signers & 157  && 20 && 21\\
         total \#unique glosses & 969 && 671 && 620 \\
         \bottomrule
    \end{tabular}
    }
    \mbox{}\vspace{-0.3cm} \\
    \caption{\textbf{Statistics} for our signer-independent split of a subset of \bslcorpus{} for which dense gloss annotations are available.
    } 
    \mbox{}\vspace{-0.6cm} \\
    \label{tab:bslcp}
\end{table}

\vspace{-0.2cm}
\subsection{Datasets and evaluation metrics}
\label{subsec:datasets}

\noindent\textbf{\bslcorpus}~\cite{bslcorpus17,schembri2013building}
is a BSL linguistic corpus that
provides various types of manual annotations, of which
approximately 72K are gloss annotations,
i.e., individual signs with their sign categories and
temporal boundaries.
We use the subset of videos where such gloss annotation
is available, and we split it into
train/validation/test
sets as in Tab.~\ref{tab:bslcp}.
The gloss annotations contain separate labels for the
right and left hand, which we merge with priority for the dominant hand.
We define sign categories according to several rules, e.g., assigning lexical variants of the same word to one class,
filtering classes with less than 10 occurrences.
Resulting annotations are used to cut the video into
shorter clips with at least three consecutive signs.
We use this data for training and evaluation.\\
\noindent\textbf{\bslonek}~\cite{albanie20_bsl1k}
is a recently collected large-scale dataset of BSL signs
for which sparse annotations are obtained using a mouthing-based
visual keyword spotting model~\cite{momeni2020seeing,stafylakis2018zero}.
This dataset does not provide precise start/end times of signs,
but rather an approximate position of the sign.
We run our segmentation model to complement the dataset
with automatic temporal annotations, which we quantitatively
evaluate on a small manually annotated subset. \\
\noindent\textbf{RWTH-PHOENIX-Weather-2014}~\cite{Koller15cslr}
(PHOENIX14)
is a standard benchmark in computer vision with
dense gloss annotations without timings for German Sign Language
(DGS) signs.
The work of~\cite{koller2017re} provides automatically generated alignments for the training
set, making use of ground-truth gloss information,
against which we compare our boundary estimations.
Since automatic timing annotations are not available
on the validation or test sets, we randomly partition
the training set into training and test splits following
a 4:1 ratio.

\noindent\textbf{Evaluation metrics.}
To assess the sign segmentation performance, we measure both the capability to predict the position of the boundary
and the extent of the sign segments.
We define a boundary prediction to be correct, if its distance to a ground-truth boundary is lower than a given threshold. The distance is measured from the mean position of the predicted and the ground-truth boundary, wherein a boundary refers to a series of contiguous 1s in $\mathbf{y}$. We calculate the F1 score for the boundaries  as the harmonic
mean of precision and recall, given this definition of a
correct boundary detection.
We use all integer-valued thresholds that fall within the closed interval [1, 4] and report the mean across thresholds, which we refer to as mF1B.
The quality of the sign segments is also evaluated with the F1 score, where sign segments with an IoU higher
than a given threshold are defined as correct. We average the results for thresholds from 0.4 to 0.75 with a step size of 0.05 and report the result as mF1S. We conduct each experiment with three different random seeds and report the mean and standard deviation.
For further interpretability of the results, we provide additional
metrics, such as the mean boundary width,
on our project page~\cite{projectpage}.


\vspace{-.2cm}
\subsection{Ablation study} \label{subsec:ablation}

We first perform several ablations on \bslcorpus{}.
\begin{table}
    \centering
    \resizebox{0.99\linewidth}{!}{
    \begin{tabular}{lrr}
         \toprule
         & mF1B & mF1S  \\
         \midrule
         Uniform baseline (using GT \#signs)  & 41.37 &  34.89 \\
         \midrule
         \bslonek{}~\cite{albanie20_bsl1k} & $58.48_{\pm1.4}$ & $33.66_{\pm2.3}$\\ 
         \bslcorpus{} (class-labels) &  $\mathbf{68.68_{\pm0.6}}$ & $\mathbf{47.71_{\pm0.8}}$ \\
         \bslonek{}$\rightarrow$ \bslcorpus{} (class-labels) & $66.17_{\pm0.5}$ & $44.44_{\pm1.0}$\\ 
         \bslonek{}$\rightarrow$ \bslcorpus{} (class-agnostic) &  $68.23_{\pm1.8}$ & $44.36_{\pm3.3}$ \\
         \bottomrule
    \end{tabular}
    }
    \mbox{}\vspace{-0.3cm} \\
    \caption{
    \textbf{The influence of I3D training data:}
    We observe that finetuning the \bslonek{}
    classification model~\cite{albanie20_bsl1k}
    on \bslcorpus{} brings a significant boost in
    performance.  However, the proposed method
    does not require category information on
    \bslcorpus{} to be effective---class-agnostic
    training suffices. 
    }
    \mbox{}\vspace{-0.6cm} \\
    \label{tab:input}
\end{table}

\noindent\textbf{Uniform baseline}.
We measure the performance
for a simple baseline
that uniformly splits a video into temporal segments given
the true number of signs.
Note that automatically counting the signs occurring in a video
is an unsolved problem, therefore this baseline does \textit{not}
represent a lower-bound, but is instead used to provide
indicative results for the metrics.
As can be seen in Tab.~\ref{tab:input}, such  a uniform
baseline is suboptimal (41.37 mF1B) even if it uses
ground-truth information.

\noindent\textbf{Sign recognition pretraining on the target data.}
Next, we compare the performance when using
three different versions of input features for MS-TCN training: pretraining the I3D model (i) on \bslonek{} 
(ii) on \bslcorpus{}
and
(iii) on \bslonek{} and finetuning on \bslcorpus{}.
In all cases, I3D is initially pretrained on Kinetics~\cite{kay2017kinetics}.
For (iii), we consider two variants: (1) finetuning using semantic \textit{class-labels} for recognition and (2) \textit{class-agnostic} finetuning in which the model is trained directly for class-agnostic boundary classification (and does not make use of the sign labels themselves). Tab.~\ref{tab:input} summarises the results
of retraining MS-TCN with each of these I3D
input features. Strikingly, while there is a clear
benefit to finetuning on \bslcorpus{}, the model
does not require the pretraining step on \bslonek{}.
Furthermore, class labels for the I3D training are not essential to achieving good segmentation performance.

\begin{figure}
    \centering
        \centering        
        \includegraphics[width=0.99\linewidth,trim=0 0 0 0.4cm, 
        clip]{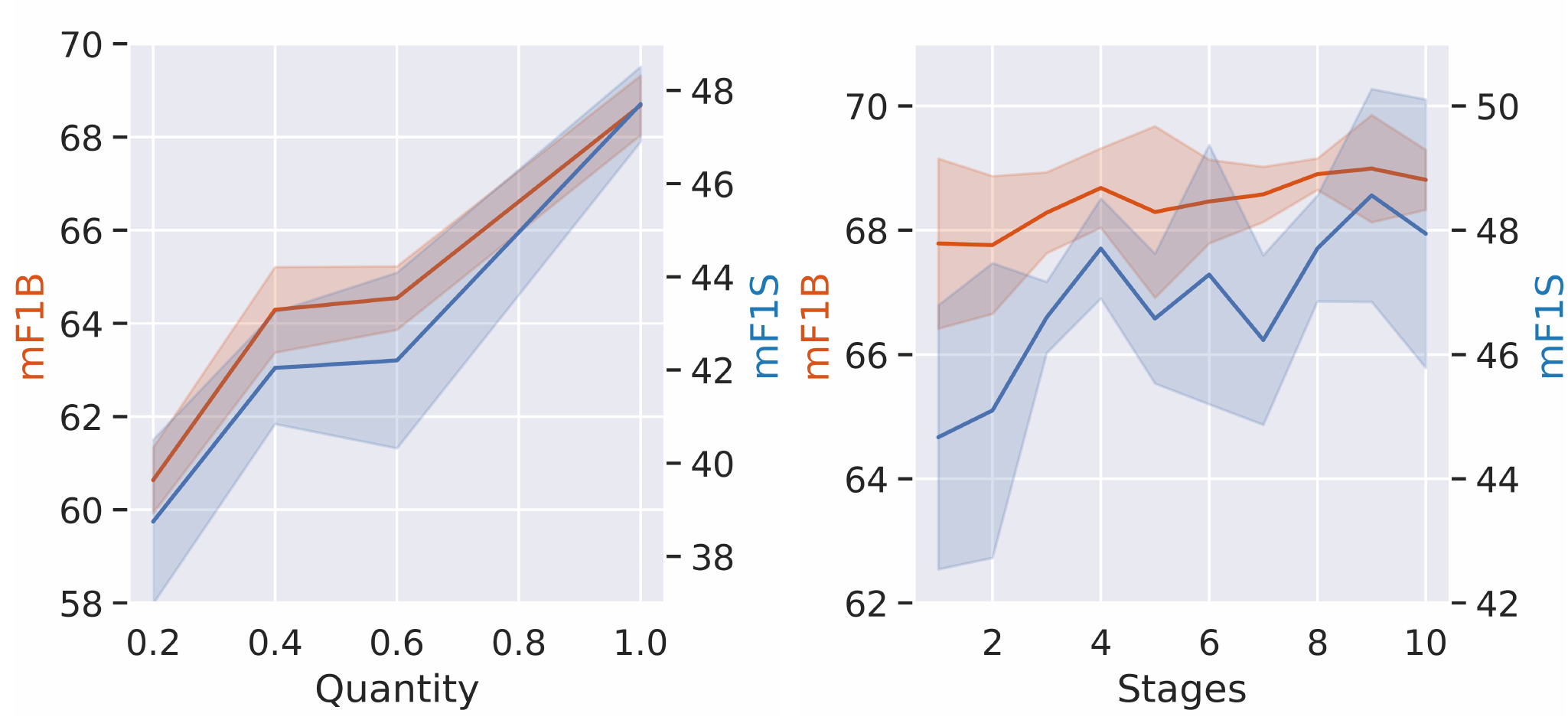}
        \centering
    \mbox{}\vspace{-0.3cm} \\
    \caption{\textbf{Ablations:} We study (left) the quantity of training data 
    available to the model,
    and (right) the number of refinement stages in the MS-TCN architecture.
    }
    \label{fig:trainingratio}
    \mbox{}\vspace{-0.6cm} \\
\end{figure}

\noindent\textbf{Quantity of training data.}
To investigate whether the quantity of available
training data represents a limiting factor for model
performance, we plot segmentation metrics
against the quantity of data used for training
both the I3D and the MS-TCN models.
As can be seen in Fig.~\ref{fig:trainingratio}, 
training data availability represents a significant bottleneck.
Exploring the use of automatically segmented
videos, which is made possible with our approach,
could be a way to mitigate the lack of training data.
We leave this to future work.

\noindent\textbf{Number of refinement stages.}
We next ablate the MS-TCN architecture by
changing the number of refinement stages.
The results in Fig.~\ref{fig:trainingratio} suggest
that one and two stages
exhibit inferior performance, likely due to lacking sufficient
access to context.
The model has diminishing benefits
from the addition of a large number of stages.
In the rest of our experiments, we employ a total of four stages as in~\cite{Farha_2019_CVPR}.


\vspace{-0.2cm}
\subsection{Comparison to prior work}
\label{subsec:sota}

In this section, we compare our model with the
method introduced in \cite{farag2019learning}, 
which uses hand-crafted geometric features
computed on 3D body pose keypoints in combination
with a Random Forest 
classifier. Since the DJSLC dataset~\cite{Brock2018DeepJA}
used by~\cite{farag2019learning} is not publicly
available to facilitate a comparison, we turn to
the publicly available \bslcorpus{} with
boundary annotations and compare to our
re-implementation of their approach. In contrast
to~\cite{farag2019learning}, which assumes 3D skeletal
information given by motion capture, we estimate the 
3D pose coordinates with the recently proposed monocular
DOPE model~\cite{dope}. While the performance is 
bounded by the quality of the pose estimation,
this ensures that the method is applicable to
unconstrained sign language videos.
For the geometric features of~\cite{farag2019learning},
we calculate angular and distance measurements for
different joint pair combinations. We compute the
Laplacian kernel matrix as described in~\cite{farag2019learning},
and concatenate the flattened upper triangle of this matrix
with the raw geometric features for a given window.
We refer the reader to~\cite{farag2019learning}
for further details.
To identify the respective influence of the features
and the classifiers, in Tab.~\ref{tab:sota}, we report
the performance of
the geometric features~\cite{farag2019learning}
with both Random Forest and MS-TCN classifiers.
A first improvement over \cite{farag2019learning}
can be attributed to the use of the MS-TCN model.
We further significantly improve
over the geometric features with our I3D finetuning on \bslcorpus{}.

\begin{table}
    \centering
    \resizebox{0.95\linewidth}{!}{
    \begin{tabular}{lcc}
         \toprule
         Method & mF1B & mF1S \\
         \midrule
         Geometric features + RF~\cite{farag2019learning} & $50.49_{\pm0.1}$& $37.46_{\pm0.1}$ \\
         \midrule
         Geometric features + MS-TCN & $60.77_{\pm2.3}$ & $36.25_{\pm2.0}$ \\
         I3D + MS-TCN (proposed) & $\mathbf{68.68_{\pm0.6}}$ & $\mathbf{47.71_{\pm0.8}}$   \\
         \bottomrule
    \end{tabular}
    }
    \mbox{}\vspace{-0.2cm} \\
    \caption{\textbf{State of the art comparison:} We compare to the geometric features proposed by~\cite{farag2019learning} on \mbox{\textbf{\bslcorpus{}}}. We show that our I3D features, combined with the MS-TCN segmentation model, significantly outperforms \cite{farag2019learning} which uses Random Forest (RF) classifiers.}
    \label{tab:sota}
\end{table}

\vspace{-0.2cm}
\subsection{Generalisation to other sign language datasets}
\label{subsec:annot}
Here, we evaluate the capability of our method to generalise on different datasets.

\noindent\textbf{\bslonek{}.} We apply our segmentation model on the \bslonek{} dataset
and complement the sparse sign annotations provided by~\cite{albanie20_bsl1k}
with timing information for start and end frames. While both our training
data \bslcorpus{} and test data \bslonek{} include the same sign language,
there is a considerable domain gap in appearance as shown in Fig.~\ref{fig:datasets}. To enable
a quantitative performance measure, we manually annotated
a 2-minute sequence with exhaustive segmentation labels,
resulting in 177 sign segments.
In Tab.~\ref{tab:bsl1k}, we observe a trend consistent with the previous experiments:
the proposed approach outperforms the prior work of~\cite{farag2019learning} by a sizeable margin. 

\begin{table}
    \centering
    \resizebox{0.9\linewidth}{!}{
    \begin{tabular}{lccc}
         \toprule
          Method & mF1B & mF1S  \\
         \midrule
         Geometric features + RF~\cite{farag2019learning} & $51.26_{\pm0.5}$ & $34.28_{\pm1.0}$\\
        I3D + MS-TCN & $\mathbf{61.12_{\pm0.9}}$ & $\mathbf{49.96_{\pm0.6}}$ \\ 
         \bottomrule
    \end{tabular}
    }
    \mbox{}\vspace{-0.2cm} \\
    \caption{\textbf{Generalisation to \bslonek{}:}
    We report the results of applying the \bslcorpus{}-trained
    segmentation model on a small fraction of \bslonek{}
    which we manually annotated.}
    \mbox{}\vspace{-0.6cm} \\
    \label{tab:bsl1k}
\end{table}

\noindent\textbf{PHOENIX14.} Next, we test the limits of our approach
on 
German Sign Language (DGS).
Note that the timing annotations are noisy due to automatic
forced-alignment~\cite{koller2017re}. Due to the challenging
domain gap both in the visual appearance and in the sign languages (BSL and DGS),
we obtain a limited but reasonable cross-lingual generalisation performance (see Tab.~\ref{tab:phoenix}),
suggesting that the model learns to exploit some common visual cues for segmenting the different sign languages. These common visual cues contain for example significant changes in direction or orientation of the hands. 
\begin{table}[t]
    \centering
    \resizebox{0.99\linewidth}{!}{
    \begin{tabular}{llcc}
         \toprule
         I3D training data & MS-TCN training data & mF1B& mF1S \\
         \midrule
          \bslcorpus{} & \bslcorpus{} & $46.75_{\pm1.2}$ & $32.29_{\pm0.3}$ \\ 
           \bslcorpus{} & PHOENIX14 & $65.06_{\pm0.5}$ & $44.42_{\pm2.0}$ \\ 
           PHOENIX14 & PHOENIX14 & $71.50_{\pm0.2}$ & $52.78_{\pm1.6}$ \\   
         \bottomrule
    \end{tabular}
    }
    \mbox{}\vspace{-0.2cm} \\
    \caption{\textbf{Generalisation to PHOENIX14 German Sign Language:} We evaluate our method
    against the automatic labels provided in PHOENIX14. The model which is only trained on \bslcorpus{} shows
    limited generalisation (46.75 mF1B) to a different sign language.
    The control experiment provides an approximate upper bound by training both the I3D and MS-TCN models using the PHOENIX14 automatic labels (71.50 mF1B) .
    }
    \mbox{}\vspace{-0.6cm} \\
    \label{tab:phoenix}
\end{table}
\begin{figure}
      \centering
      \includegraphics[width=0.48\textwidth]{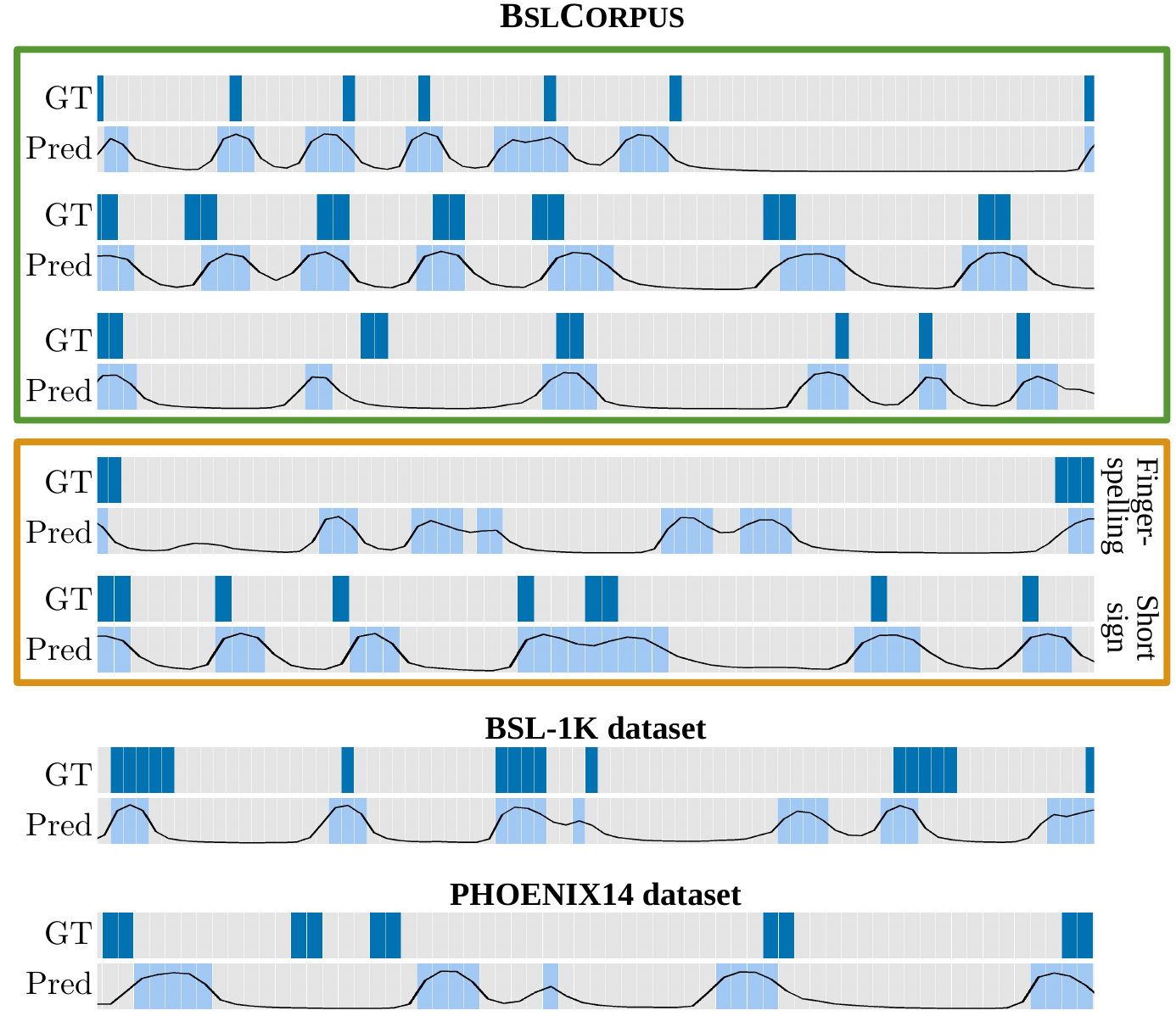}
      \mbox{}\vspace{-0.6cm} \\
      \caption{\textbf{Qualitative analysis:} Results on \bslcorpus{} (top),
      \bslonek{} and PHOENIX14 (bottom) datasets. We illustrate success
      (green) and failure cases (orange) on \bslcorpus{}, with the latter 
      arising from fingerspelling and a very short sign. }
      \mbox{}\vspace{-0.6cm} \\
      \label{fig:qual}
\end{figure}

\vspace{-0.3cm}
\subsection{Qualitative analysis}
\label{subsec:qual}
In Fig.~\ref{fig:qual}, we provide qualitative results on all three datasets: \bslcorpus{}, \bslonek{}, and PHOENIX14.
The examples enclosed in the orange box show
two different failure categories on \bslcorpus{}.
In one case, our model over-segments a
fingerspelled word by predicting the boundaries
of individual letters.
We also observe that the model has difficulty with
very short signs.
For more qualitative analysis and the
corresponding videos, we refer the reader to our project page~\cite{projectpage}.

\vspace{-0.4cm}
\section{Conclusion} \label{sec:conclusion}

In this paper we addressed the problem of temporal sign language segmentation. 
We employed temporal convolutions and formulated the problem as sign boundary detection. We provided a comprehensive study to analyse our various components. We further reported the results on three datasets
with varying properties showcasing our method's generalisation capabilities. Future directions include extending
the amount of training data by exploring automatic annotations.

{
\vspace{0.2cm}
\small
\noindent\textbf{Acknowledgements:}
This work was supported by EPSRC grant ExTol. KR was supported by the German Academic Scholarship Foundation. The authors would like to express their gratitude to C. Camgoz for the help with the \bslcorpus{} data preparation, N. Fox for his invaluable assistance in providing an annotation sequence for evaluation, L. Momeni and A. Braffort for valuable feedback. 
SA would like to acknowledge the support of Z. Novak and S. Carlson in enabling his contribution to this research.
}


\clearpage

{
\small
\bibliographystyle{IEEEtranS}
\setlength{\bibsep}{1.9pt plus 0.1ex}
\bibliography{refs}

\begin{thebibliography}{10}
\providecommand{\url}[1]{#1}
\csname url@samestyle\endcsname
\providecommand{\newblock}{\relax}
\providecommand{\bibinfo}[2]{#2}
\providecommand{\BIBentrySTDinterwordspacing}{\spaceskip=0pt\relax}
\providecommand{\BIBentryALTinterwordstretchfactor}{4}
\providecommand{\BIBentryALTinterwordspacing}{\spaceskip=\fontdimen2\font plus
\BIBentryALTinterwordstretchfactor\fontdimen3\font minus
  \fontdimen4\font\relax}
\providecommand{\BIBforeignlanguage}[2]{{%
\expandafter\ifx\csname l@#1\endcsname\relax
\typeout{** WARNING: IEEEtranS.bst: No hyphenation pattern has been}%
\typeout{** loaded for the language `#1'. Using the pattern for}%
\typeout{** the default language instead.}%
\else
\language=\csname l@#1\endcsname
\fi
#2}}
\providecommand{\BIBdecl}{\relax}
\BIBdecl

\bibitem{albanie20_bsl1k}
S.~Albanie, G.~Varol, L.~Momeni, T.~Afouras, J.~S. Chung, N.~Fox, and
  A.~Zisserman, ``{BSL-1K}: {S}caling up co-articulated sign language
  recognition using mouthing cues,'' in \emph{ECCV}, 2020.

\bibitem{braffort2012}
A.~Braffort and L.~Boutora, ``{DEGELS}2012 annotation challenge: Segmentation
  [in {F}rench],'' in \emph{JEP-TALN-RECITAL, DEGELS: Gestures and Sign
  Language Challenge)}, 2012.

\bibitem{bragg2019sign}
D.~Bragg, O.~Koller \emph{et~al.}, ``Sign language recognition, generation, and
  translation: An interdisciplinary perspective,'' in \emph{ACM SIGACCESS},
  2019.

\bibitem{brentari2009effects}
D.~Brentari, ``Effects of language modality on word segmentation: An
  experimental study of phonological factors in a sign language,'' \emph{Papers
  in laboratory phonology, vol.8}, pp. 155--164, 2009.

\bibitem{Brock2018DeepJA}
H.~Brock and K.~Nakadai, ``Deep {JSLC}: A multimodal corpus collection for
  data-driven generation of {Japanese Sign Language} expressions,'' in
  \emph{LREC}, 2018.

\bibitem{bull2020}
H.~Bull, M.~Gouiff\`es, and A.~Braffort, ``Automatic segmentation of sign
  language into subtitle-units,'' in \emph{ECCVW (SLRTP)}, 2020.

\bibitem{carreira2017quo}
J.~Carreira and A.~Zisserman, ``Quo vadis, action recognition? {A} new model
  and the kinetics dataset,'' in \emph{CVPR}, 2017.

\bibitem{Cooper2009LearningSF}
H.~Cooper and R.~Bowden, ``Learning signs from subtitles: A weakly supervised
  approach to sign language recognition,'' \emph{CVPR}, 2009.

\bibitem{cormierbsl}
K.~Cormier and J.~Fenlon, ``{BSL} corpus annotation guidelines,'' 2014.

\bibitem{casborn2008}
O.~Crasborn and I.~Zwitserlood, ``Annotation of the video data in the corpus
  {NGT},'' \emph{Radboud University Nijmegen}, 2008.

\bibitem{Dreuw2008TowardsAS}
P.~Dreuw and H.~Ney, ``Towards automatic sign language annotation for the
  {ELAN} tool,'' in \emph{LREC Workshop on the Representation and Processing of
  Sign Languages: Construction and Exploitation of Sign Language Corpora},
  2008.

\bibitem{farag2019learning}
I.~Farag and H.~Brock, ``Learning motion disfluencies for automatic sign
  language segmentation,'' in \emph{ICASSP}, 2019.

\bibitem{Farha_2019_CVPR}
Y.~A. Farha and J.~Gall, ``{MS-TCN}: Multi-stage temporal convolutional network
  for action segmentation,'' in \emph{CVPR}, 2019.

\bibitem{gonzalez}
M.~Gonzalez, ``Computer vision methods for unconstrained gesture recognition in
  the context of sign language annotation,'' Ph.D. dissertation, Universite de
  Toulouse, 2012.

\bibitem{hanke2012}
T.~Hanke, S.~Matthes, A.~Regen, and S.~Worseck, ``Where does a sign start and
  end? segmentation of continuous signing,'' in \emph{LREC Workshop on the
  Representation and Processing of Sign Languages: Interactions between Corpus
  and Lexicon}, 2012.

\bibitem{Joze19msasl}
H.~R.~V. Joze and O.~Koller, ``{MS-ASL}: {A} large-scale data set and benchmark
  for understanding american sign language,'' in \emph{BMVC}, 2019.

\bibitem{kay2017kinetics}
W.~Kay, J.~Carreira, K.~Simonyan, B.~Zhang, C.~Hillier, S.~Vijayanarasimhan,
  F.~Viola, T.~Green, T.~Back, P.~Natsev, M.~Suleyman, and A.~Zisserman, ``The
  {Kinetics} human action video dataset,'' \emph{arXiv}, 2017.

\bibitem{KollerPami20}
O.~{Koller}, N.~C. {Camgoz}, H.~{Ney}, and R.~{Bowden}, ``Weakly supervised
  learning with multi-stream {CNN-LSTM-HMMs} to discover sequential parallelism
  in sign language videos,'' \emph{IEEE Transactions on Pattern Analysis and
  Machine Intelligence}, 2020.

\bibitem{koller2020quantitative}
O.~Koller, ``Quantitative survey of the state of the art in sign language
  recognition,'' \emph{arXiv:2008.09918}, 2020.

\bibitem{Koller15cslr}
O.~Koller, J.~Forster, and H.~Ney, ``Continuous sign language recognition:
  Towards large vocabulary statistical recognition systems handling multiple
  signers,'' \emph{Computer Vision and Image Understanding}, vol. 141, 2015.

\bibitem{koller2017re}
O.~Koller, S.~Zargaran, and H.~Ney, ``Re-sign: Re-aligned end-to-end sequence
  modelling with deep recurrent {CNN-HMMs},'' in \emph{CVPR}, 2017.

\bibitem{li2020word}
D.~Li, C.~Rodriguez, X.~Yu, and H.~Li, ``Word-level deep sign language
  recognition from video: A new large-scale dataset and methods comparison,''
  in \emph{WACV}, 2020.

\bibitem{momeni2020seeing}
L.~Momeni, T.~Afouras, T.~Stafylakis, S.~Albanie, and A.~Zisserman, ``Seeing
  wake words: Audio-visual keyword spotting,'' \emph{BMVC}, 2020.

\bibitem{momeni20watchread}
L.~Momeni, G.~Varol, S.~Albanie, T.~Afouras, and A.~Zisserman, ``Watch, read
  and lookup: learning to spot signs from multiple supervisors,'' in
  \emph{ACCV}, 2020.

\bibitem{moryossef2020real}
A.~Moryossef, I.~Tsochantaridis, R.~Aharoni, S.~Ebling, and S.~Narayanan,
  ``{Real-Time Sign Language Detection using Human Pose Estimation},'' in
  \emph{ECCVW, Sign Language Recognition, Translation and Production (SLRTP)},
  2020.

\bibitem{Santemiz2009}
P.~{Santemiz}, O.~{Aran}, M.~{Saraclar}, and L.~{Akarun}, ``Automatic sign
  segmentation from continuous signing via multiple sequence alignment,'' in
  \emph{ICCVW}, 2009.

\bibitem{bslcorpus17}
\BIBentryALTinterwordspacing
A.~Schembri, J.~Fenlon, R.~Rentelis, and K.~Cormier, ``{British Sign Language
  Corpus Project: A corpus of digital video data and annotations of British
  Sign Language 2008-2017 (Third Edition)},'' 2017. [Online]. Available:
  \url{http://www.bslcorpusproject.org}
\BIBentrySTDinterwordspacing

\bibitem{schembri2013building}
A.~Schembri, J.~Fenlon, R.~Rentelis, S.~Reynolds, and K.~Cormier, ``Building
  the {B}ritish {S}ign {L}anguage {C}orpus,'' \emph{Language Documentation \&
  Conservation}, vol.~7, pp. 136--154, 2013.

\bibitem{shao2020finegym}
D.~Shao, Y.~Zhao, B.~Dai, and D.~Lin, ``{FineGym}: A hierarchical video dataset
  for fine-grained action understanding,'' in \emph{CVPR}, 2020.

\bibitem{projectpage}
``Sign segmentation project page,''
  \url{https://www.robots.ox.ac.uk/~vgg/research/signsegmentation/}.

\bibitem{stafylakis2018zero}
T.~Stafylakis and G.~Tzimiropoulos, ``Zero-shot keyword spotting for visual
  speech recognition in-the-wild,'' in \emph{ECCV}, 2018.

\bibitem{sutton-spence_woll_1999}
R.~Sutton-Spence and B.~Woll, \emph{The Linguistics of {B}ritish Sign Language:
  An Introduction}.\hskip 1em plus 0.5em minus 0.4em\relax Cambridge University
  Press, 1999.

\bibitem{theodorakis2014dynamic}
S.~Theodorakis, V.~Pitsikalis, and P.~Maragos, ``Dynamic--static unsupervised
  sequentiality, statistical subunits and lexicon for sign language
  recognition,'' \emph{Image and Vision Computing}, vol.~32, no.~8, pp.
  533--549, 2014.

\bibitem{vaswani2017attention}
A.~Vaswani, N.~Shazeer, N.~Parmar, J.~Uszkoreit, L.~Jones, A.~N. Gomez,
  {\L}.~Kaiser, and I.~Polosukhin, ``Attention is all you need,'' in
  \emph{NeurIPS}, 2017.

\bibitem{dope}
P.~Weinzaepfel, R.~Br\'egier, H.~Combaluzier, V.~Leroy, and G.~Rogez, ``{DOPE}:
  Distillation of part experts for whole-body {3D} pose estimation in the
  wild,'' in \emph{ECCV}, 2020.

\end{thebibliography}
}

\end{document}